\documentclass[runningheads]{llncs}

\usepackage[T1]{fontenc}

\usepackage{graphicx,verbatim}
\usepackage{multirow}
\usepackage{enumitem}

\usepackage{amsmath}
\usepackage{cite}
\usepackage{amssymb}

\usepackage{hyperref}

\begin{document}

\title{MLRecon: Robust Markerless Freehand 3D Ultrasound Reconstruction via Coarse-to-Fine Pose Estimation}
\titlerunning{MLRecon: Robust Markerless Freehand 3D Ultrasound Reconstruction}

\author{Yi Zhang$^1$, Puxun Tu$^1$, Kun Wang$^1$, Yulin Yan$^2$, Tao Ying$^2$, Xiaojun Chen$^{1,3}$}  
\authorrunning{Yi Zhang et al.}
\institute{$^1$ Institute of Biomedical Manufacturing and Life Quality Engineering, School of Mechanical Engineering, Shanghai Jiao Tong University, Shanghai, China \\
		$^2$ Department of Ultrasound in Medicine, Shanghai Sixth People's Hospital Affiliated to Shanghai Jiao Tong University School of Medicine, Shanghai, China  \\
		$^3$ Institute of Medical Robotics, Shanghai Jiao Tong University, Shanghai, China}

\maketitle             

\begin{abstract}

Freehand 3D ultrasound (US) reconstruction promises volumetric imaging with the flexibility of standard 2D probes, yet existing tracking paradigms face a restrictive trilemma: marker-based systems demand prohibitive costs, inside-out methods require intrusive sensor attachment, and sensorless approaches suffer from severe cumulative drift. To overcome these limitations, we present \textbf{MLRecon}, a robust markerless 3D US reconstruction framework delivering drift-resilient 6D probe pose tracking using a single commodity RGB-D camera. Leveraging the generalization power of vision foundation models, our pipeline enables continuous markerless tracking of the probe, augmented by a vision-guided divergence detector that autonomously monitors tracking integrity and triggers failure recovery to ensure uninterrupted scanning. Crucially, we further propose a dual-stage pose refinement network that explicitly disentangles high-frequency jitter from low-frequency bias, effectively denoising the trajectory while maintaining the kinematic fidelity of operator maneuvers. Experiments demonstrate that MLRecon significantly outperforms competing sensorless and sensor-aided methods, achieving average position errors as low as 0.88 mm on complex trajectories and yielding high-quality 3D reconstructions with sub-millimeter mean surface accuracy. This establishes a new benchmark for low-cost, accessible volumetric US imaging in resource-limited clinical settings.


\end{abstract}

\section{Introduction}

\begin{figure}
	\includegraphics[width=\textwidth]{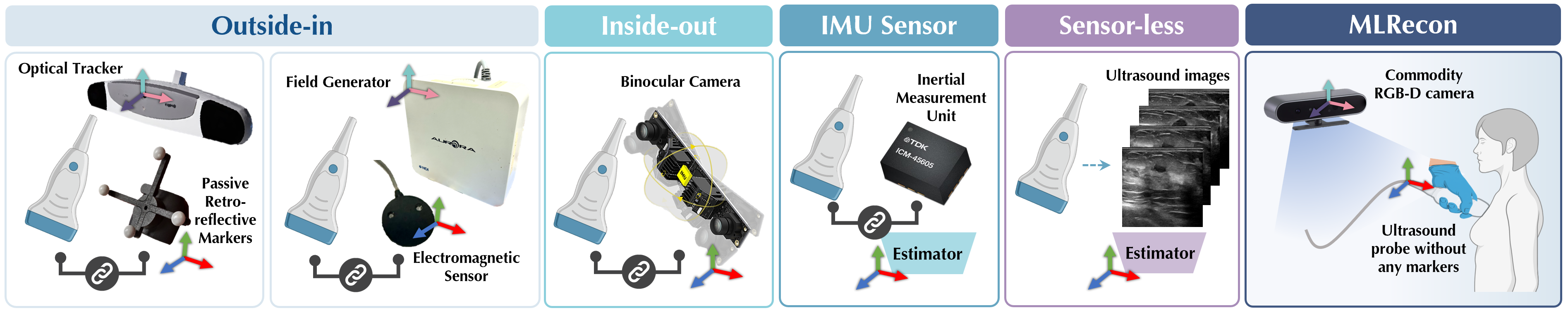}
	\caption{Schematic comparison of tracking paradigms in freehand 3D ultrasound.} \label{fig1}
\end{figure}

Ultrasound (US) imaging is indispensable in image-guided interventions owing to its real-time capability, portability, and low cost. However, conventional 2D acquisition suffers from high operator dependency and lacks spatial awareness. 3D US has therefore attracted growing interest for providing richer anatomical context that facilitates volumetric assessment, procedural planning, and diagnosis~\cite{fenster2001three, mozaffari2017freehand}. Existing clinical solutions typically rely on dedicated 3D probes or mechanically swept transducers, which require specialized hardware and restrict the scanning range. In contrast, freehand 3D US reconstruction offers an attractive alternative by continuously tracking the 6D pose of a standard 2D probe and compounding the B-mode frames into a coherent volumetric representation, thereby preserving the flexibility and accessibility of handheld ultrasonography.

Accurate and robust probe pose tracking is fundamental to high-fidelity freehand US reconstruction, and a wide spectrum of solutions has been explored (Fig.~\ref{fig1}). \emph{Marker-based outside-in} systems, including the optical~\cite{treece2003high} and electromagnetic (EM) ~\cite{daoud2015freehand} trackers, deliver high precision but incur substantial cost and depend on dedicated infrastructures. To eliminate external trackers, \emph{inside-out} paradigms mount lightweight sensors directly on the probe. 
Prominent inertial-centric investigations range from standalone Inertial Measurement Unit (IMU) approaches~\cite{rahni20082d} to deep IMU+US fusion networks~\cite{luo2022deep, yan2024fine, luo2025monetv2} for pose estimation. Incorporating visual exteroception,
Buchanan~et~al.~\cite{buchanan20243d} demonstrated that visual-inertial odometry from a probe-mounted stereo camera achieves reconstruction errors comparable to motion capture, while He~et~al.~\cite{he2024freehand} fused a binocular camera with an IMU via an unscented Kalman filter on Lie groups. Zhang~et~al.~\cite{zhang2025freehand} further introduced a simulation-in-the-loop visual servoing scheme with probe-mounted monocular cameras observing a textured planar workspace. Although these inside-out methods obviate the need for external tracking equipment, they necessitate physical sensor attachment to the probe and remain susceptible to incremental drift over extended trajectories.

At the other extreme, \emph{sensorless} approaches predict inter-frame relative poses purely from US image sequences using deep learning~\cite{li2023long, luo2023recon, wilson2025dualtrack, lee2025enhancing}, thereby avoiding any auxiliary hardware. However, these methods suffer from substantial cumulative drift, particularly over clinically realistic non-linear scanning paths, and their reliance on learned speckle or texture patterns inherently limits generalization across different anatomical sites and US machines. Recently, Dai~et~al.~\cite{dai2024advancing} mitigated drift by embedding N-shaped nylon fiducials inside a hydrogel coupling pad that imprints spatial markers into US images, yet the pad physically constrains the scanning surface and is restricted to linear probes.

An ideal freehand 3D US tracking solution for clinical intervention guidance should be (i)~\emph{markerless}, requiring no modifications to the probe or attachments on the patient, (ii)~\emph{drift-resilient} over arbitrarily long and complex trajectories, and (iii)~\emph{lightweight and low-cost} for seamless integration into existing workflows.
None of the aforementioned paradigms simultaneously satisfies all three criteria.

In this study, we present \textbf{MLRecon}, a completely markerless freehand 3D US reconstruction framework that bridges this gap. Our main contributions are twofold: (i) A foundation-model-based probe pose estimation and tracking pipeline with a vision-guided divergence detector that enables automatic failure recovery, providing uninterrupted tracking throughout freehand scanning. (ii) A dual-stage temporal pose refinement network that separately decouples and removes high-frequency jitter and low-frequency bias, substantially reducing maximum pose deviations while faithfully preserving genuine probe dynamics.

\section{Method}
\label{sec:method}

\begin{figure}
	\includegraphics[width=\textwidth]{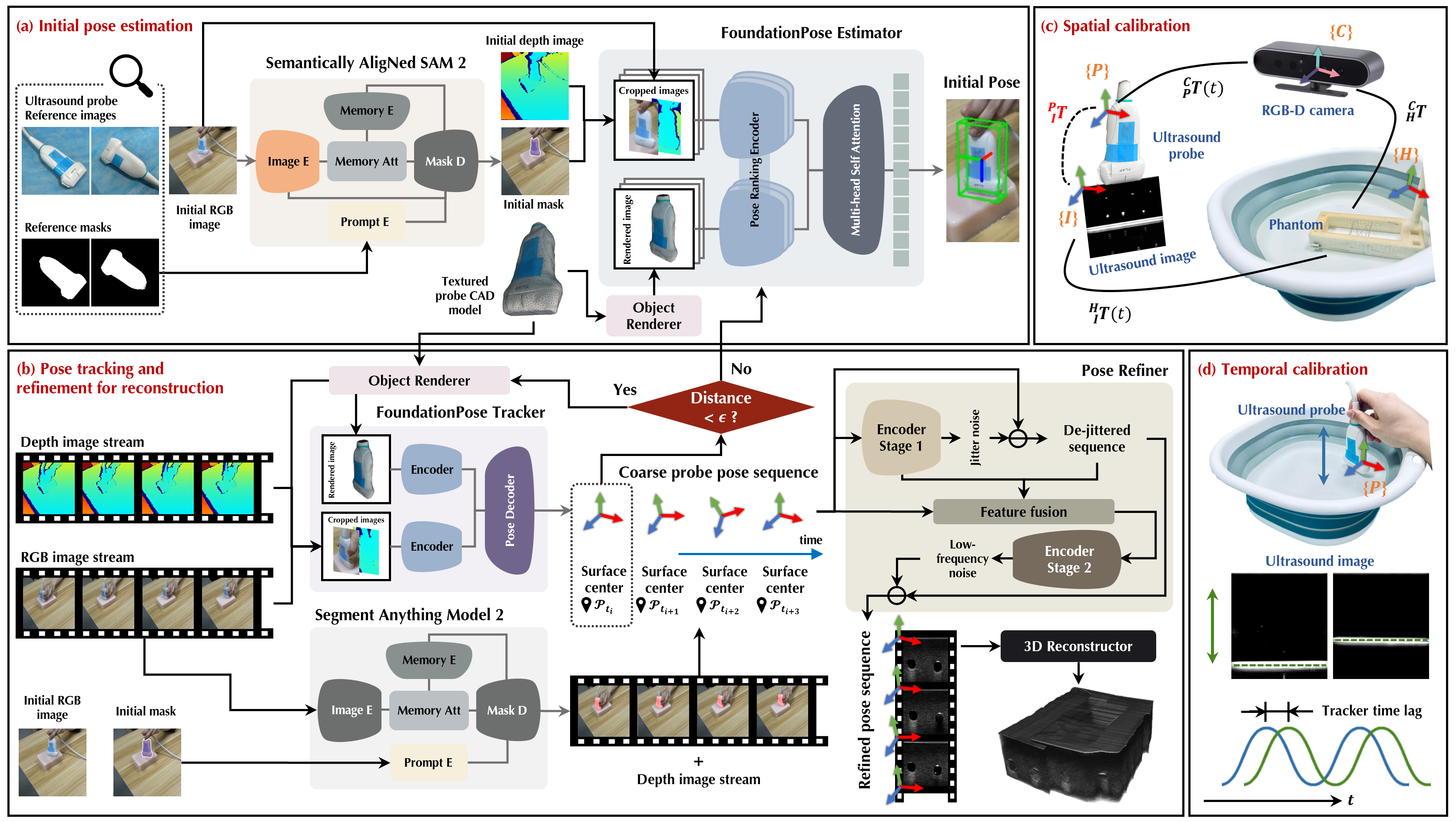}
	\caption{Overview of MLRecon. The framework integrates robust tracking with failure recovery, dual-stage frequency-aware pose refinement, and calibrated 3D compounding. } \label{fig2}
\end{figure}

As illustrated in Fig.~\ref{fig2}, we employ a single commodity RGB-D camera (Astra\,2, Orbbec Inc., 30\,Hz) as the external tracking device, observing a clinical US probe (Mindray 7L4P, 5.0--10.0\,MHz) during scanning. The proposed pipeline incorporates a cascade of foundation-model-based modules for robust probe pose estimation and tracking. We denote the coordinate frames of the RGB-D camera, US probe, and US image as $\{C\}$, $\{P\}$, and $\{I\}$, respectively.

\subsection{Foundation-Model-Based Pose Estimation and Tracking}
\label{subsec:pose_est_track}

To achieve markerless tracking for freehand reconstruction, our approach estimates the 6D probe pose $_{P}^{C}\hat{\mathbf{T}} \in \mathrm{SE}(3)$ directly from the RGB-D stream using a render-and-compare paradigm built on FoundationPose~\cite{wen2024foundationpose}. Two critical challenges must be addressed: obtaining a reliable initial pose without manual annotation, along with detecting and recovering from inevitable tracking failures.

\textbf{Initial pose estimation.} Given $K$ annotated reference images $\{(\mathbf{I}_r^k, \mathbf{a}_r^k)\}_{k=1}^{K}$ of the probe from diverse viewpoints, we employ SANSA~\cite{cuttano2025sansa}, a semantically aligned adaptation of SAM\,2~\cite{ravi2024sam2}, to propagate object masks to the live first RGB frame $\mathbf{I}_{r,t_0}$ without manual annotation. The predicted mask $\hat{\mathbf{S}}_{t_0}$, the initial RGB-D observation, and a pre-scanned CAD model $\mathcal{M}$ of the probe are fed to FoundationPose, which performs global pose sampling followed by iterative render-and-compare refinement to yield the initial pose $_{P}^{C}\hat{\mathbf{T}}_{t_0}$ (Fig.~\ref{fig2}a).

\textbf{Robust tracking with divergence detection and recovery.}
Once initialized, FoundationPose tracks the probe pose frame by frame on the incoming RGB-D stream at 30 Hz. However, visual tracking of rigid objects is susceptible to transient failures due to extensive occlusions, depth sensor noise, or abrupt probe motions. We therefore introduce a \emph{vision-guided tracking divergence detector} that monitors the consistency between the tracked pose and an independent visual cue, triggering automatic re-initialization upon disagreement.

Concurrently with FoundationPose tracking, we run SAM\,2 at a reduced cadence ($\sim$3\,Hz) to segment the probe in the current RGB frame, producing a binary mask $\hat{\mathbf{S}}_{t_i}$. From this mask and the aligned depth map $\mathbf{D}_{t_i}$, we compute a visual centroid $\mathbf{c}^{\mathrm{vis}}_{t_i}$ by back-projecting the masked depth pixels into 3D:
\begin{equation}
	\mathbf{c}^{\mathrm{vis}}_{t_i} 
	= \frac{1}{|\mathcal{P}|}
	\sum_{(u,v)\in \mathcal{P}}
	\mathbf{D}_{t_i}(u,v)\,\mathbf{K}^{-1}\!\begin{pmatrix}u\\v\\1\end{pmatrix},
	\label{eq:visual_centroid}
\end{equation}
where $\mathcal{P}=\{(u,v)\mid\hat{\mathbf{S}}_{t_i}(u,v)>0,\;\mathbf{D}_{t_i}(u,v)\in[D_{\min},D_{\max}]\}$ is the set of valid masked pixels and $\mathbf{K}$ is the depth camera intrinsic matrix. 
Simultaneously, the tracked probe centroid $\mathbf{c}^{\mathrm{trk}}_{t_i}$ is derived from $_{P}^{C}\hat{\mathbf{T}}_{t_i}$ at the corresponding timestamp.

Tracking divergence is declared when the Euclidean discrepancy exceeds an adaptive, object-aware threshold:
\begin{equation}
	\lVert \mathbf{c}^{\mathrm{vis}}_{t_i} - \mathbf{c}^{\mathrm{trk}}_{t_i} \rVert_2 
	\;>\;\epsilon\;=\; 
	\frac{\lVert\mathbf{d}_{\mathcal{M}}\rVert_2}{2}\,\eta
	\;+\; 
	\delta_0,
	\label{eq:divergence}
\end{equation}
where $\mathbf{d}_{\mathcal{M}}\!\in\!\mathbb{R}^3$ denotes the bounding-box dimensions of $\mathcal{M}$, $\eta$ is a geometric scaling factor, and $\delta_0$ is a fixed tolerance. Upon detection at $t_k$, 
the tracker is halted; the current SAM\,2 mask and RGB-D frame are leveraged to re-initialize the pose, after which tracking seamlessly resumes. Poses between $t_k$ and the previous check are marked as missing. This closed-loop mechanism operates transparently to the sonographer, requiring no manual intervention during scanning.

\subsection{Dual-Stage Pose Refinement}
\label{subsec:pose_refine}

Even after robust tracking, the coarse pose sequence $_{P}^{C}\hat{\boldsymbol{\mathcal{T}}}=\{_{P}^{C}\hat{\mathbf{T}}_{t_1},\dots,_{P}^{C}\hat{\mathbf{T}}_{t_L}\}$ inevitably exhibits two distinct types of error that degrade reconstruction quality: (i)~\emph{high-frequency jitter} stemming from per-frame depth noise and independent network predictions; (ii)~\emph{low-frequency deviations} primarily arising from auto-regressive initialization that propagates under-corrected residuals. Classical filters address only one regime, risking either over-smoothing genuine motion or leaving correlated bias unresolved.

To address this, we propose \emph{Pose Refiner}, a convolutional temporal network that explicitly disentangles and eliminates both error components via a two-stage residual architecture (Fig.~\ref{fig2}b). Each pose $_{P}^{C}\hat{\mathbf{T}}_{t_i}$ is parameterized as $\mathbf{x}_{t_i}=[\mathbf{r}_{t_i};\,\mathbf{p}_{t_i}] \in \mathbb{R}^{9}$, where $\mathbf{r}\!\in\!\mathbb{R}^6$ is the continuous rotation representation~\cite{zhou2019continuity} and $\mathbf{p}\!\in\!\mathbb{R}^3$ is the translation. 
The raw sequence is normalized via an affine map $\tilde{\mathbf{X}} = \boldsymbol{\Lambda}^{-1}(\mathbf{X} - \bar{\mathbf{x}}_{\mathrm{ref}})$ with $\boldsymbol{\Lambda}=\mathrm{diag}(\mathbf{I}_6,\,s_p\mathbf{I}_3)$ and offset $\bar{\mathbf{x}}_{\mathrm{ref}}=[\mathbf{0}_6;\,\mathbf{p}_{\mathrm{ref}}]$, where $\mathbf{p}_{\mathrm{ref}}$ is the first-frame position and $s_p$ is a spatial scaling factor, yielding $\tilde{\mathbf{X}}\!\in\!\mathbb{R}^{9\times L}$.

\textbf{Stage\,1} employs a dilated temporal convolutional encoder $\mathcal{E}_1$
with restricted dilations $\{1,2,4,8,16\}$ to isolate local jitter patterns:
\begin{equation}
	\hat{\mathbf{n}}^{\mathrm{hf}} = \mathcal{E}_1(\tilde{\mathbf{X}}),
	\quad
	\tilde{\mathbf{X}}^{(1)} = \tilde{\mathbf{X}} - \hat{\mathbf{n}}^{\mathrm{hf}}.
	\label{eq:stage1}
\end{equation}
A lightweight fusion layer then concatenates $\tilde{\mathbf{X}}$, the de-jittered signal $\tilde{\mathbf{X}}^{(1)}$, and the Stage\,1 semantic features to form $\mathbf{F}_{\mathrm{fuse}}$, supplying complementary multi-scale cues to the subsequent stage.


\textbf{Stage\,2} utilizes another encoder $\mathcal{E}_2$ with aggressively increased dilations $\{1,2,\dots,128\}$
whose receptive field spans the entire sequence, 
enabling it to capture the residual low-frequency bias. The refined output is recovered by the symmetric denormalization:
\begin{equation}
	\hat{\mathbf{n}}^{\mathrm{lf}} = \mathcal{E}_2(\mathbf{F}_{\mathrm{fuse}}),
	\quad
	\mathbf{X}^{\star}
	= \boldsymbol{\Lambda}
	\bigl(\tilde{\mathbf{X}}^{(1)} - \hat{\mathbf{n}}^{\mathrm{lf}}\bigr)
	+ \bar{\mathbf{x}}_{\mathrm{ref}}.
	\label{eq:stage2}
\end{equation}

\textbf{Training objective.}
The network is trained on 243 scans ($\sim$360k frames) collected at the Department of Ultrasound in Medicine, Shanghai Sixth People's Hospital, pairing simulated noisy RGB-D poses with clean optical tracking sequences.
The composite loss jointly supervises $\mathbf{X}^{\star}$ and ${\mathbf{X}}^{(1)}$:
\begin{equation}
	\mathcal{L}_{\mathrm{comp}}
	= \lambda_g \mathcal{L}_{\mathrm{geo}}
	+ \lambda_l \mathcal{L}_{\ell_1}
	+ \lambda_v \mathcal{L}_{\mathrm{vel}}
	+ \lambda_f \mathcal{L}_{\mathrm{freq}},
	\label{eq:loss}
\end{equation}
where $\mathcal{L}_{\mathrm{geo}}$ is the geodesic distance on $\mathrm{SO}(3)$ \cite{salehi2018real}, $\mathcal{L}_{\ell_1}$ is the L1 error for rotation and translation representations, $\mathcal{L}_{\mathrm{vel}}$ penalizes temporal-derivative discrepancies to preserve motion dynamics, and $\mathcal{L}_{\mathrm{freq}}$ enforces spectral magnitude agreement via the real FFT to prevent over-smoothing of genuine motion frequencies.

\subsection{Calibration and 3D Ultrasound Compounding}
\label{subsec:recon}

To map each B-mode pixel into 3D space, we determine the spatial transform $_{I}^{P}{\mathbf{T}}$ from the US image frame $\{I\}$ to the probe frame $\{P\}$ via an improved N-wire phantom calibration protocol~\cite{carbajal2013improving} (Fig.~\ref{fig2}c). Temporal calibration aligns the US and camera streams by maximizing the cross-correlation of quasi-periodic reciprocating motions~\cite{treece2003high} (Fig.~\ref{fig2}d). For 3D compounding, each B-mode pixel $p{=}(u,v)$ is mapped to physical coordinates $\mathbf{x}_{t_i} = {_{P}^{C}\hat{\mathbf{T}}}^{\star}_{t_i}\,{_{I}^{P}{\mathbf{T}}}\,(u,v,0,1)^T$---where ${_{P}^{C}\hat{\mathbf{T}}}^{\star}_{t_i}$ denotes the refined pose from Sec.~\ref{subsec:pose_refine}---and assigned to the nearest voxel via bin-filling, with empty voxels inpainted by gradient-aware hole-filling~\cite{wen2013accurate}.

\section{Experiments and Results}

\begin{figure}
	\centering
	\includegraphics[width=\textwidth]{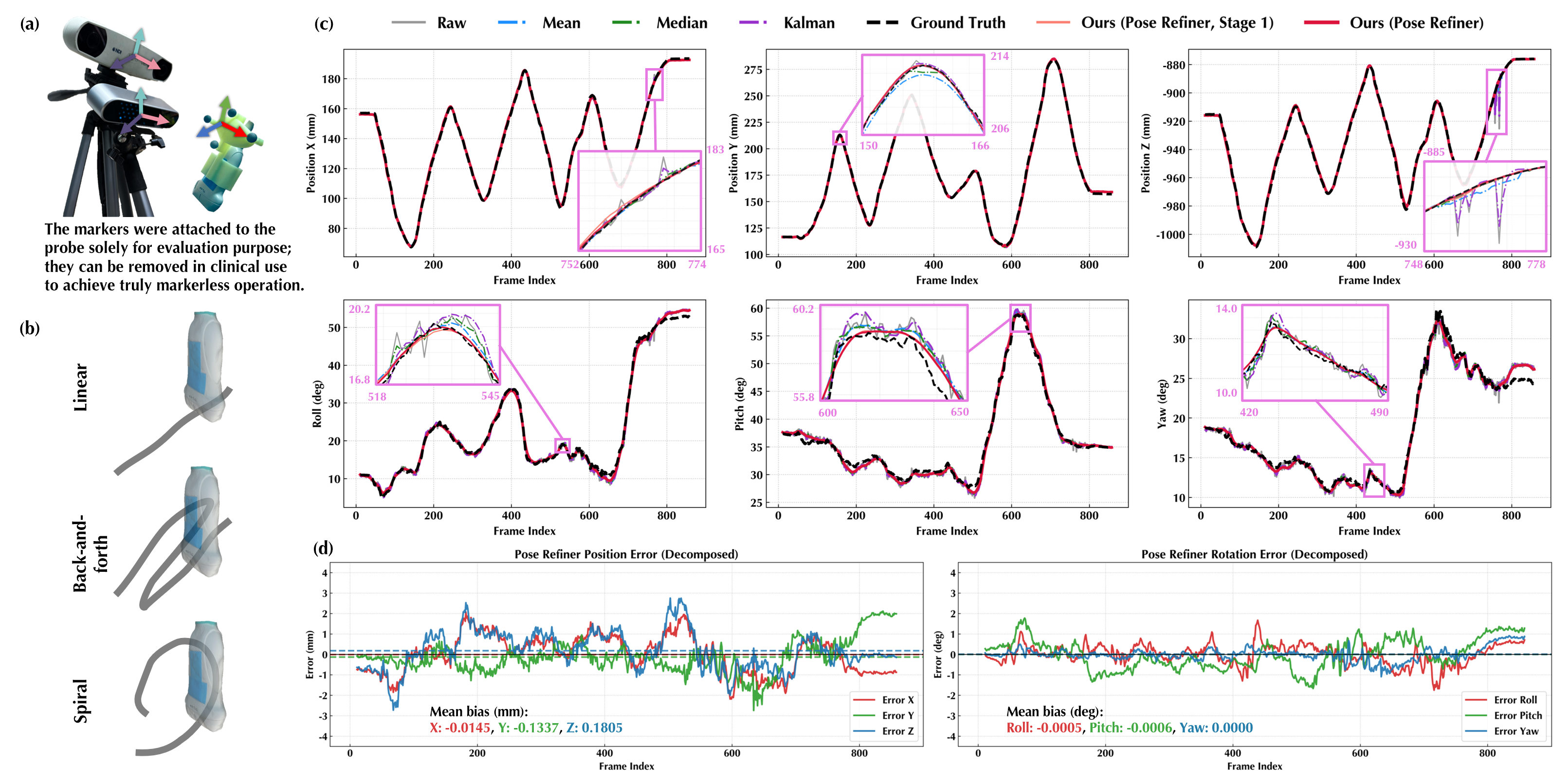}
	\caption{(a) Experimental setup. (b) Three scanning trajectory modes. (c) Representative translation (X/Y/Z) and rotation (Roll/Pitch/Yaw) time-series from the ablation study. (d) Per-Degree-of-Freedom (DOF) pose error after dual-stage pose refinement.} \label{fig3}
\end{figure}

In our implementation, MLRecon has been integrated into an existing US data capture and reconstruction software~\cite{zhang2026navigation}, where FoundationPose and SAM 2 were deployed using C++ TensorRT-accelerated pipelines on an NVIDIA RTX 3080 Ti laptop GPU, supporting
tracking at up to 140\,Hz. US image–pose pairs were acquired at 20\,Hz for 3D compounding. Reference images for SANSA initialization were collected from $K{=}2$ viewpoints.
For divergence detection, we set $\eta{=}0.8$ and $\delta_0{=}30$\,mm. The detector triggered successful re-initialization in all 10 dedicated trials involving transient complete invisibility or rapid motion of the probe, demonstrating the reliability of the recovery mechanism.
The proposed Pose Refiner was trained with AdamW (learning rate$\,{=}\,10^{-3}$, cosine annealing) for 1000\,epochs with batch size 32 
($s_p{=}100$, $\lambda_g{=}1.0$, $\lambda_l{=}5.0$, $\lambda_v{=}3.0$, $\lambda_f{=}0.1$).
The experimental setup for pose estimation and reconstruction is shown in Fig.~\ref{fig3}a.

\textbf{Pose accuracy.} We evaluated MLRecon across three representative freehand trajectories (Fig.~\ref{fig3}b) and compared against existing methods. Metrics include Final Drift Rate (FDR), Average Drift Rate (ADR), Maximum Drift (MD), Maximum Rotation Error (MRE), Average Position Error (APE), Average Rotation Error (ARE), and Total Distance (TD). Sensorless and IMU+US fusion methods \cite{luo2023recon, dai2024advancing, luo2022deep} were compared on a linear sweep, while other sensor-aided methods were compared on back-and-forth and spiral trajectories with comparable ranges \cite{buchanan20243d, he2024freehand, busam2018markerless, zhang2025freehand}. All MLRecon results were averaged over 5 trials.

\begin{table}[]
	\caption{The mean(std) pose accuracy results across three scanning trajectory modes.}
	\centering
	\label{tab:1}
	\setlength{\tabcolsep}{0.3pt}
	\small
	\begin{tabular}{ccccccc}
		\hline
		\textbf{Mode}                  & \textbf{Methods}    & \textbf{FDR(\%)}       & \textbf{ADR(\%)}       & \textbf{MD(mm)}     & \textbf{MRE(deg)}    & \textbf{TD(mm)}        \\ \hline
		\multirow{4}{*}{Linear} & RecON \cite{luo2023recon}     & 25.90(18.33) & 30.36(13.95) & --         & --         & $\sim$80   \\
		& MoNet \cite{luo2022deep}     & 12.75(9.05)  & 19.05(11.46) & --         & --         & $\sim$80   \\
		& Dai et al. \cite{dai2024advancing}    & 2.74(2.98)   & 3.35(3.24)   & 2.52(2.25) & --         & $\sim$80   \\
		& MLRecon    & \textbf{0.36(0.15)} & \textbf{0.27(0.05)} & \textbf{1.85(0.33)} & \textbf{1.12(0.14)} & $\sim$250 \\ \hline
		\hline
		\textbf{Mode}     &    \textbf{Methods}      &    \textbf{APE(mm)}    &    \textbf{ARE(deg)}   &    \textbf{MD(mm)}     & \textbf{MRE(deg)}   & \textbf{TD(mm)}        \\ \hline
		\multirow{5}{*}{\begin{tabular}[c]{@{}c@{}}Back\\  and \\ forth\end{tabular}} & Buchanan et al. \cite{buchanan20243d} & $\sim$4.00 & --         & --   & --   & $\sim$360 \\
		& He et al. \cite{he2024freehand}    & 3.78       & \textbf{0.36}       & --         & --         & $\sim$800  \\
		& ORB-SLAM  \cite{busam2018markerless}   & 2.65(0.74) & 1.99(1.99) & --         & --         & --            \\
		& Zhang et al. \cite{zhang2025freehand}   & 1.46(0.70) & 1.37(0.57) & 2.99       & 2.28       & $\sim$520  \\
		& MLRecon      & \textbf{0.88(0.45)} & 0.48(0.23) & \textbf{2.37(0.25)} & \textbf{1.33(0.22)} & $\sim$560 \\ \hline
		\multirow{2}{*}{Spiral}                                                       & Zhang et al. \cite{zhang2025freehand}    & 1.98(1.43) & 1.57(0.82) & 5.72 & 3.25 & $\sim$400 \\
		& MLRecon      & \textbf{1.44(0.69)} & \textbf{0.92(0.34)} & \textbf{3.29(0.52)} & \textbf{1.75(0.15)} & $\sim$480 \\ \hline
	\end{tabular}
\end{table}

As presented in Table~\ref{tab:1}, on the linear sweep, MLRecon achieved a $7.6\times$ lower FDR and $12.4\times$ lower ADR compared to the best competing sensorless method,
despite covering a trajectory $3\times$ longer. 
Furthermore, it attained the lowest position errors on back-and-forth and spiral trajectories, surpassing all compared inside-out methods with competitive rotation accuracy.
The substantially reduced maximum deviations further demonstrate the stability of our method.

\textbf{Ablation study on pose refinement.} Table~\ref{tab:2} compares Pose Refiner against classical signal-processing baselines (Kalman, mean, and median filters) and a Stage-1-only variant, all applied to the same raw FoundationPose sequences across 5 free-scanning (random combination of the three modes) trials.

The raw sequence already achieved acceptable average errors but suffers from occasional large spikes. Both mean and Kalman filters mitigated MD but elevated average position or rotation errors due to outlier smearing and rigid statistical priors. Stage\,1 alone yielded meaningful improvement, and adding Stage\,2 further reduced MD to $3.73$\,mm with the best rotational stability (MRE $1.88$°), demonstrating that the two-stage decomposition is essential. Fig.~\ref{fig3}c--d visualize trajectory fidelity, confirming near-zero mean bias across all six DOFs.

\begin{table}[]
	\caption{The mean(std) results of ablation study on pose refinement.}
	\centering
	\label{tab:2}
	\setlength{\tabcolsep}{5.0pt}
	\small
	\begin{tabular}{ccccc}
		\hline
		& \textbf{APE(mm)}    & \textbf{ARE(deg)}   & \textbf{MD(mm)}      & \textbf{MRE(deg)}    \\ \hline
		Raw       & 1.46(0.10) & 0.75(0.09) & 25.92(9.34) & 2.63(0.40) \\
		Kalman    & 1.46(0.10) & 0.78(0.09) & 18.82(6.08) & 2.42(0.33) \\
		Mean      & 1.79(0.16) & \textbf{0.70(0.09)} & 6.05(1.05)  & 2.21(0.42) \\
		Median    & 1.45(0.06) & 0.71(0.09) & 4.91(0.69)  & 2.23(0.31) \\
		Pose Refiner (S1) & 1.41(0.07) & 0.71(0.08) & 3.90(0.24)  & 2.04(0.14) \\
		Pose Refiner (S1+S2)     & \textbf{1.39(0.06)} & 0.71(0.07) & \textbf{3.73(0.37)}  & \textbf{1.88(0.19)} \\ \hline
	\end{tabular}
\end{table}

\begin{figure}
	\centering
	\includegraphics[width=\textwidth]{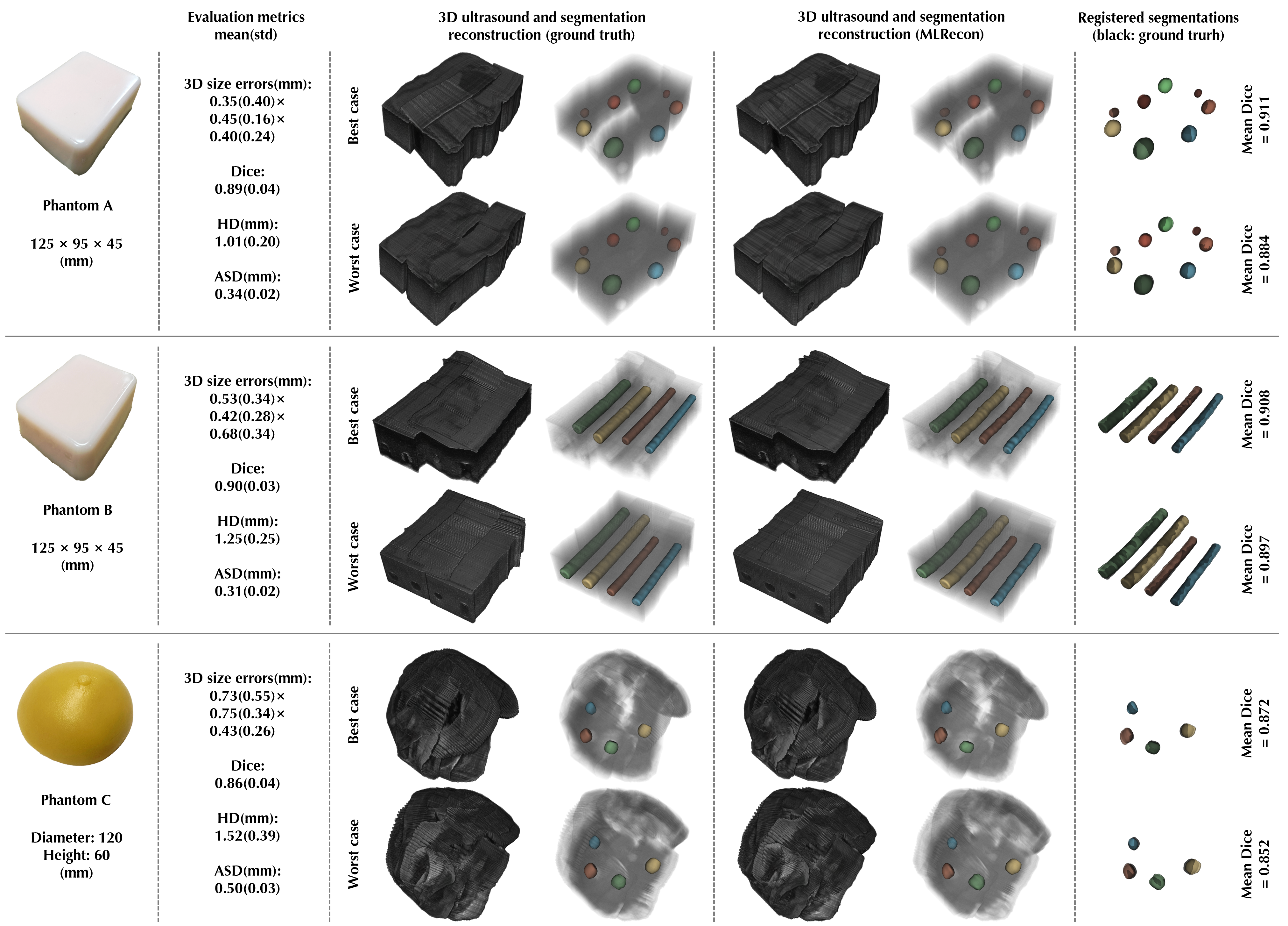}
	\caption{Quantitative and qualitative comparison of reconstruction on three phantoms.} \label{fig4}
\end{figure}

\textbf{3D reconstruction accuracy.} To evaluate the practicality of our method, we performed 3D US reconstruction on three phantoms: two rectangular tissue phantoms (Phantom A and B, containing 8 lesions and 4 vessels, respectively) and one breast-shaped phantom (Phantom C, containing 4 lesions), with 3 repeated trials per phantom. All reconstruction results were compared against ground truth volumes acquired via NDI optical tracking, following the registration-based protocol detailed in~\cite{zhang2026navigation}. Metrics include 3D size error along principal axes, Dice coefficient, Hausdorff Distance (HD), and Average Surface Distance (ASD) between registered lesions or vessels.

As shown in Fig.~\ref{fig4}, the reconstructed 3D volumes and segmentations from our method closely aligned with ground truth morphology in both best- and worst-cases, with Dice coefficients ranging from 0.85 to 0.91 across all phantoms. Even for the uneven-surfaced Phantom~C, sub-millimeter accuracy was maintained in both 3D size error and ASD, underscoring that scanning different body geometries exerts minimal influence on reconstruction quality.

\section{Conclusion}

In this study, we present MLRecon, a markerless freehand 3D US reconstruction framework that incorporates vision foundation models and a single commodity RGB-D camera. By combining closed-loop divergence detection and frequency-aware pose refinement, the system attains state-of-the-art precision on complex trajectories and delivers high-fidelity volumetric representations across multiple phantoms. Crucially, its marker-free, calibration-light workflow facilitates seamless clinical integration without modifying routine scanning practices, positioning it as a robust and cost-effective alternative to optical or EM tracking with significant promise for enabling accessible point-of-care diagnostics.

%
%

\bibliographystyle{splncs04}
\bibliography{refs}

\end{document}